\documentclass{article}

 \usepackage[preprint]{neurips_2026}

\usepackage[utf8]{inputenc} 
\usepackage[most]{tcolorbox}
\usepackage[T1]{fontenc}    
\usepackage{hyperref}       
\usepackage{url}            
\usepackage{booktabs}       
\usepackage{multirow}
\usepackage{amsfonts}       
\usepackage{nicefrac}       
\usepackage{microtype}      
\usepackage{xcolor}         
\usepackage{graphicx}
\usepackage{subcaption}
\usepackage{placeins}   
\definecolor{darkblue}{rgb}{0, 0, 0.5}
\hypersetup{colorlinks=true, citecolor=darkblue, linkcolor=darkblue, urlcolor=darkblue}
\definecolor{promptcolor}{RGB}{245, 245, 250} 
\newtcolorbox{promptbox}{
  colback=promptcolor,
  colframe=blue!50!black,
  boxrule=0.5pt,
  arc=2pt,
  left=4pt,
  right=4pt,
  top=4pt,
  bottom=4pt,
  width=\columnwidth, 
  before skip=8pt,
  after skip=8pt,
}

\title{IDEAFix: Evaluation Framework for Creative Defixation Prompting in LLMs}

\author{
Florian Carichon$^{1,2}$\thanks{Corresponding author} \and
Soumya Sharma$^{1,2}$ \and
Meaghan Girard$^{2}$ \and
Romain Rampa$^{3}$ \and
Golnoosh Farnadi$^{1,2}$ \\
\\
$^{1}$ McGill University, Montreal, Canada \\
$^{2}$ Mila, Montreal, Canada \\
$^{3}$ Concordia University, Montreal, Canada \\
$^{4}$ ÉTS, Montreal, Canada \\
\\
\texttt{\{florian.carichon,soumya.sharma,farnadig\} @mila.quebec},  \\
\texttt{meaghan.girard@concordia.ca},
\texttt{romain.rampa@etsmtl.ca}
}

\begin{document}

\maketitle

\begin{abstract}

  Large language models (LLMs) are increasingly used for tasks involving creative problem solving and idea generation. However, there is a lack of consensus concerning their creative capabilities: some studies report superior performances compared to humans, while others highlight structural limitations such as fixation and the homogenization of outputs. Existing evaluation approaches either rely on narrow, decontextualized tasks that do not capture goal-oriented generation or on broader settings that confound multiple aspects of the creative process, making it difficult to isolate the effects of task formulation, prompting, and evaluation design. Significantly, the role of structured prompting strategies in shaping idea generation remains underexplored. Therefore, we introduce IDEAFix, an evaluation framework for analyzing divergent thinking in open-ended idea generation tasks. We prompt models to generate multiple original solutions to controlled variations of short design scenarios, task attributes, and defixation prompting strategies. This design enables systematic analysis of how structured guidance influences LLMs' idea generation. Our results show that both task formulation and attribute selection significantly affect models' performance, and that simple prompting strategies can boost the originality of solutions. However, we also observe persistent output homogenization across models, confirming inherent limits in their ability to generate diverse solutions. Overall, IDEAFix provides a controlled, extensible framework for studying the mechanisms underlying LLMs' creativity.\footnote{\href{https://github.com/soummyaah/CreativityAndEthicalFramework}{Code} and \href{https://huggingface.co/datasets/soummyaah/IDEAFix}{Dataset} released here.}
\end{abstract}

\section{Introduction}
Large Language Models (LLMs) are increasingly deployed in tasks that require creative capabilities, including story-writing \citep{herbold2023large}, product design ideation \citep{meincke2024using}, or scientific discovery \citep{reddy2025towards}. As a result, assessing the creative abilities of these systems has become an important challenge. Creativity is typically defined as the capacity to produce ideas that are novel, useful, and surprising \citep{boden2004creative}. In studies of human creativity, problem-solving is defined as the mental process of searching for solutions to a given problem or objective. This process relies on divergent thinking, which is the ability to expand from an initial concept into a space of ideas and identify those that are both original and relevant \citep{ismayilzada2024creativity}.

Divergent thinking has become standard for evaluating LLMs' creative capabilities. Many existing evaluation approaches are adapted from classical psychological tests, such as the Alternative Uses Task (AUT) \citep{li2025automated}, the Torrance Tests of Creative Thinking (TTCT) \citep{rabeyah2024llms}, or the Divergent Association Task (DAT) \citep{bellemare2024divergent}. Across these settings, a consistent finding is that LLMs can generate ideas that are perceived as both novel and useful, and in some contexts may match or exceed average human performance \citep{conroy2024ai,hubert2024current}. However, other studies on idea generation tasks report a tendency toward homogenization in LLM outputs, with models producing highly convergent responses across prompts and contexts \citep{jiang2025artificial,anderson2024homogenization}. Finally, researchers have made similar observations in human-AI creativity, where individual performance improves but collective diversity decreases \citep{doshi2024generative}.

Beyond divergent thinking, the nature of LLM creativity remains an open question. Prior works suggest that LLMs primarily generate creative outputs by recombining and extending existing knowledge, and are fundamentally unable to redefine the underlying space of possible ideas \citep{franceschelli2025creativity} due to their training paradigms, which make them reliant on patterns learned from observed data \citep{west2025base}. Moreover, LLMs' creativity may be overestimated: apparent diversity often reflects superficial variation \citep{peeperkorn2024temperature}, and divergent thinking evaluations emphasize associative and analogy tasks that favor semantic pattern matching over creative reasoning \citep{liu2026parallelograms}. Therefore, researchers rightfully raise the question of whether LLMs outperform humans due to genuine creative abilities, or because they rely on semantic manipulation and data contamination \citep{bellemare2024divergent}, especially when models struggle with unconventional and unseen ideas \citep{huang2024lateval,tian2024macgyver}. 
 
An explanation for this paradox is fixation, where ideation is constrained by familiar patterns and prior knowledge \citep{agogue2014impact,wadinambiarachchi2024effects}. Emerging evidence suggests that LLMs exhibit fixation effects, overproducing conventional ideas and fewer original ones, with this tendency increasing as task complexity grows \citep{cheng2025inspiration,desdevises2025paradox}. Research in creativity management and design has led to the development of numerous methodologies aimed at reducing these fixations, such as brainstorming, C-K, TRIZ, and Design Thinking \citep{hatchuel2004ck,moehrle2005triz,liedtka2018design}. Since prompting strategies significantly influence LLM performance \citep{liu2023pre, wei2022chain}, LLM creativity may be sensitive to how the generation process is guided, motivating the use of structured prompts inspired by these established creativity or defixation methods.

In this article, we introduce IDEAFix, an evaluation framework for open-ended idea generation and divergent thinking in LLMs under controlled conditions. It combines variations of adjective attributes (e.g., positive, negative, surprising, traditional) with briefs describing products, services, strategies, or procedures. This design enables systematic and flexible generation of ideation scenarios. IDEAFix also incorporates structured prompting strategies inspired by creativity and defixation methods. Generated ideas are evaluated automatically along standard divergent thinking dimensions. By jointly varying task structure and prompting, IDEAFix provides a controlled framework to assess both LLMs' creative ability for idea generation and their sensitivity to enhancing strategies. Our contributions are summarized as follows:

\begin{itemize}
    \item  We introduce IDEAFix, an evaluation framework for open-ended idea generation and divergent thinking in LLMs.
    \item  We construct a dataset combining diverse scenario briefs and attribute expansions, along with a structured prompting protocol inspired by creativity and defixation methods.
    \item We conduct a systematic analysis across multiple LLMs. Specifically, we examine how task structures and prompting strategies affect individual divergent idea generation and the emergence of collective homogenization in generated outputs.
\end{itemize}

\section{Related Section}

\subsection{Evaluation approaches for divergent thinking in LLMs}
A growing body of work evaluates LLM creativity across multiple processes and task types such as divergent and convergent thinking, elaboration and problem solving, narrative generation, and scientific discovery \citep{ismayilzada2024creativity}. Most studies on divergent thinking rely on classic tasks such as AUT, TTCT, and DAT \citep{brown2020language,goes2023pushing,guzik2023originality, bellemare2024divergent}. Across all these studies, a common conclusion is that LLMs are perceived as highly creative, sometimes matching or exceeding average human performance \citep{hubert2024current}. However, these evaluation approaches share a key limitation: creativity is often operationalized as generating numerous, semantically distant ideas under minimal constraints. While this effectively captures diversity, it provides only a limited assessment of originality or contextual utility. In that regard, other lines of work explore alternative settings, such as lateral thinking \citep{huang2024lateval}, open-ended problem solving \citep{tian2024macgyver}, product improvement and improbable consequences \citep{anderson2024homogenization}, or diversity in real-world queries \citep{jiang2025artificial}.  However, these approaches often mix multiple aspects of the creative process, including tasks, evaluated concepts, or the role of prompting strategies, when considered. As a result, it becomes difficult to determine which aspects of LLMs' divergent thinking are evaluated, since idea generation and evaluation are treated solely as output properties rather than as a guided process \citep{cheng2025inspiration}. While prior work shows that prompting can enhance idea generation in human-AI co-creation settings \citep{haase2025has}, it remains limited to only a few prompts for the AUT and DAT tasks. In contrast, we introduce a controlled evaluation framework that enables systematic analysis of how  structured prompting strategies and task design influence divergent idea generation in LLMs.

\subsection{Creativity methods as structured prompting strategies}
One approach to steering LLMs' performances is through structured prompting strategies \citep{wei2022chain,liu2023pre}. In human creativity research and design practice, a wide range of techniques has been developed to guide ideation processes and help humans go beyond conventional ideas. Classical methods include brainstorming \citep{putman2009brainstorming}, design thinking \citep{liedtka2018design}, TRIZ \citep{moehrle2005triz}, C-K theory \citep{hatchuel2004ck}, SCAMPER \citep{ozyaprak2016effectiveness}, and scenario-based forecasting \citep{de2000brief}. Although these approaches differ in their implementation, they share a common objective: shaping how individuals search through a space of possible ideas. Recent work has explored adapting creativity methods to generative AI \citep{cheng2025inspiration}. For instance, design thinking phases have been used as prompting strategies to produce more user-focused and iterative outputs \citep{bockle2026design}. At the same time, methods such as SCAMPER and TRIZ have been integrated into prompts to guide exploration of the solution space \citep{chulvi2025design, jiang2025autotriz}. However, these studies remain limited in scope and lack systematic evaluation, often considering only a single model, a small set of prompting strategies, or a specific ideation task. As a result, it motivates the need for a benchmark that enables systematic evaluation of how structured prompting strategies inspired by methods that are already proven to effectively stimulate human creativity can effectively influence LLMs' idea generation.

\section{IDEAFix Evaluation Framework}
\label{sec:protocol}
IDEAFix is an evaluation framework that follows a 4-step process to evaluate the creative capacity of LLMs. Figure \ref{fig:IDEAFix_process} presents the IDEAFix pipeline, and an overview of the dataset card is provided in appendix \ref{appendix:dataset}, including its intended use and associated risks.

\begin{figure}[h]
    \centering
    \includegraphics[width=0.7\textwidth]{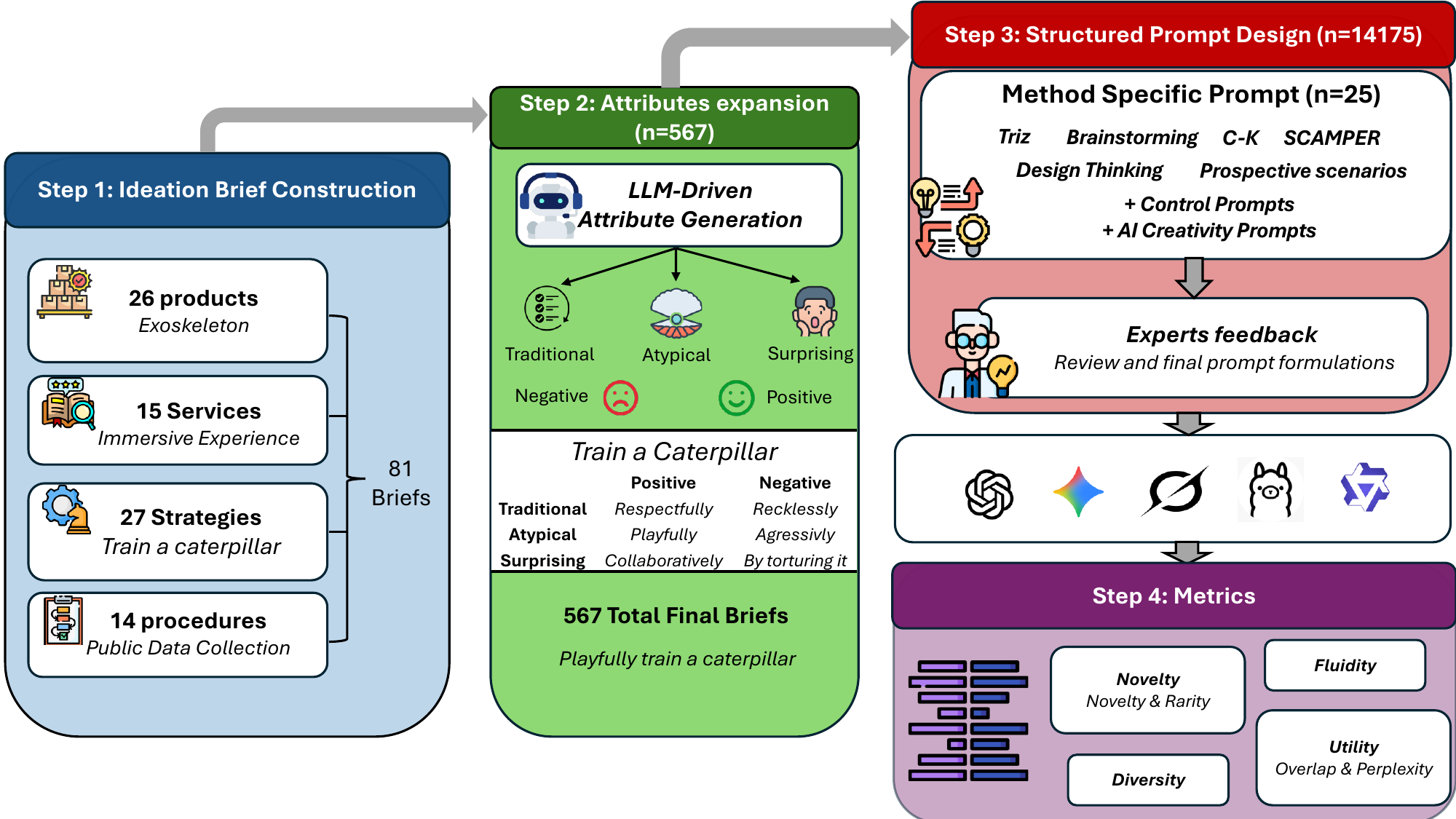}
    \caption{IDEAFix protocol depicting the 3 steps to create the dataset (Brief Construction, Attributes Expansion, Prompt Design) and the 4th step  to evaluate generated solutions.}
    \label{fig:IDEAFix_process}
\end{figure}

\subsection{Brief construction}
In designing IDEAFix, we first aimed to construct a diverse and representative corpus of ideation briefs spanning products, services, processes, and strategies across multiple domains. To ensure a broad coverage of real-world problem types, the dataset includes standard design tasks and more complex scenarios, such as ethical dilemmas and sensitive topics, allowing the evaluation of model behavior under varying constraints. The briefs were collected through a combination of literature review in creativity and ethics, analysis of sensitive content categories \citep{wang2023not}, and expert-authored contributions from domain experts in ethics and innovation management. This resulted in a corpus of 81 briefs balanced across key design categories. More specifically, we created 26 briefs on products, 15 on services, 26 on strategies, and 14 on procedures. Among those, 11 are foundational creativity tasks coming from the social science literature on creativity (see Appendix \ref{sec:appendix_IDEAFix} for  references). Briefs include tasks across categories, such as product design (e.g., “irrigation system”), service design (e.g., “buying a train ticket online”), design of strategies (e.g., “messages for a political campaign”), and procedures (e.g., “promoting diversity in a university”).

\subsection{Attributes expansion}
To enrich the brief corpus and systematically investigate the effect of semantic variation on model outputs, each brief was expanded into 6 attribute-based variants. We manipulated the typicality of the adjective associated with the brief along three levels: 1) Traditional attributes (frequently associated with the brief domain); 2) Atypical attributes (rarely associated with the brief domain), 3) Surprising attributes (where the association with the brief domain is unexpected). For each brief, we used chatGPT (OpenAI, GPT-5) to generate candidate attribute words, both positive and negative in valence, at each of the three typicality levels, producing six total candidate variations per brief. Then, attribute selection was performed iteratively, retaining the most contextually relevant adjectives and ensuring syntactic consistency, both guided by expert judgment from domain experts in ethics and innovation management. In addition, a control condition was preserved for each brief (e.g., brief without any attribute), resulting in a total of 567 experimental conditions. Step 2 in Figure \ref{fig:IDEAFix_process} presents an example of attribute expansion for the brief "Train a Caterpillar". Moreover, Figure \ref{fig:IDEAFix_expansion} in Appendix \ref{sec:appendix_IDEAFix} provides additional examples of our brief expansion process.

\subsection{Prompts design}
In this part, we developed a structured set of 25 prompts. We first established 2 control prompts featuring simple semantic variations of a solution-generation template, intended to serve as baselines against which method-specific prompts could be compared. Then, we developed a set of method-specific prompts grounded in the principles and procedural steps of 6 established creativity and design methods, applied through the ideation phase. The methods include 3 Brainstorming \citep{putman2009brainstorming} formulation variants, 4 C-K Theory \citep{hatchuel2004ck} variants, 4 TRIZ \citep{moehrle2005triz} variants, 4 Design Thinking \citep{liedtka2018design} variants, 3 Prospective scenarios \citep{de2000brief} variants, and 1 Scamper \citep{ozyaprak2016effectiveness} variant. Finally, we introduce 4 AI-specific strategies designed for LLMs to study whether new defixation methodologies should be deployed to improve AI creative capabilities. These strategies are oriented around category negation, where the principle is that novelty for AI can emerge by forcing models to not explore its simplest exploratory path (e.g., generating conventional ideas) by adding simple constraints (e.g., asking LLMs to generate ideas, categorize them, and then generate new ideas outside of those  categories) \citep{lehman2008exploiting}.

To ensure cross-prompt comparability, all prompts were engineered around a common structural template comprising: (i) a concise description of the method, (ii) a step-by-step reasoning methodology \citep{wei2022chain}, (iii) the introduction of the expanded brief created in the previous step, and (iv) a set of shared output instructions, including the beginning of sentence "The solutions are" to guide the LLMs toward the desired output \citep{zhang2022automatic}. This standardized architecture ensures that observed differences in creative output can be attributed to methodological variation rather than confounds in prompt structure. Figure \ref{prompt:instance_example} presents a control prompt as well as one of our AI-specific prompt alternatives. More examples of prompting strategies are provided in Appendix \ref{sec:appendix_IDEAFixPrompts}.

\begin{figure}
\begin{promptbox}
Generate as many diverse and creative solutions as possible to address the following design challenge: [Brief context]. 
\textit{Instructions:}
List all the solutions you find as a list.
Each solution is composed of a maximum of two sentences enabling to understand the general concept behind the solution.
All solutions should be written in English.
The solutions are:
\end{promptbox}

\begin{promptbox}
Invent as many different and imaginative responses as possible to the following situation: [Brief context]. \textbf{Go beyond obvious answers — include wild, futuristic, or playful directions that might open new ways of thinking.}
\textit{Instructions:}
List all the solutions you find as a list.
Each solution is composed of a maximum of two sentences enabling to understand the general concept behind the solution.
All solutions should be written in English.
The solutions are:
\end{promptbox}
\caption{Examples of alternative Control and AI-specific prompting strategies.}
\label{prompt:instance_example}
\end{figure}

Finally, to ensure prompt validity and fidelity to the original methodologies, each prompt was reviewed by domain experts in creativity methods (including C-K theory, brainstorming, design thinking, TRIZ, and SCAMPER), all academic researchers in the field of management of innovation and creativity. These experts provided structured feedback, which was systematically reviewed and integrated into the final prompt formulations prior to the dataset deployment.

\subsection{Idea generation protocol}
\label{sec:idea_splitting}
The complete dataset consists of 14,350 input prompts, each used to generate a list of solutions from the LLMs. Since most of our analyses occur at the individual idea level, for each input prompt we process the generated output list to extract our structured idea sets. This extraction first removes any content that does not belong to a structured list; specifically, any text preceding the first list element is excluded. Moreover, when multiple lists of ideas are generated within a single output, only the final list is retained, as it most consistently reflects the model’s complete response. Then we employ a set of textual delimiters and markers, including (but not limited to) newline characters \textbackslash n, tabulations \textbackslash t, and explicit patterns such as \textit{Solution XX:} to form our structured list. To reduce noise in the output set, we perform several additional cleaning steps on the solutions themselves. First, we remove incorrect or corrupted output pairs caused by models' refusal to answer due to negatively framed keywords in certain briefs, or because responses do not conform to the expected formatting constraints. We also discard from our analyses any list with 1 or fewer ideas, as these are mainly corrupted outputs. Finally, we filter any ideas containing five words or fewer, as they are considered insufficiently developed to convey a meaningful concept. All these preprocessing rules were defined through an iterative manual inspection of model outputs, by analyzing recurring failure cases and refining the extraction procedure until stable and consistent behavior was observed.

\subsection{Metrics for divergent thinking in idea generation}

In this work, we study creativity through divergent thinking in open-ended problem solving tasks to isolate the generative phase of idea production in a controlled setting. Divergent thinking refers to the ability to generate multiple, varied, and novel ideas \citep{silvia2008assessing,haase2025has}. It is commonly evaluated through dimensions such as fluency, originality (novelty or rarity), diversity, and utility \citep{amabile1983social,silvia2008assessing}. For originality and diversity, we adopt the notion of distances within a representational idea space, in line with conceptualizations of creativity as exploration within a structured knowledge space \citep{boden2008computers}. In design and computational creativity research, these dimensions are commonly operationalized as some distance relative to a set of previously generated solutions, enabling comparisons across ideas \citep{grace2015data,fiorineschi2023uses}. In defixation studies, this reference set is commonly defined by a control group not guided by methods, examples, or defixation attributes \cite{agogue2014impact}. Finally, following common practices in NLP and computational creativity, we approximate this space using semantic distances in embedding representations of generated ideas \citep{feuerriegel2025using,jiang2025artificial}.

Therefore, we use the following metrics. \textbf{Fluency} measures the number of ideas generated for a given prompt input. \textbf{Diversity} represents the average pairwise distance of all individual ideas. \textbf{Novelty} represents the distance of an idea from a set of common previously generated ideas. This set is defined here as the ideas produced by our two control prompting strategies, i.e. the neutral brief with the two control prompt alternatives as they represent ideas produced by models without being guided by defixation methods. \textbf{Rarity} similarly measures the distance of an idea from the ensemble of ideas generated by a model across all prompting strategies for a given task. All distance-based metrics are computed in embedding space using cosine similarity, with values bounded between 0 and 1. \textbf{Overlap} captures the extent to which the brief and its associated keywords overlap in the final output, also bounded between 0 to 1. And finally, \textbf{Perplexity} measures how well a language model predicts a given sequence. 

All mathematical and implementation details for the metrics are provided in Appendix \ref{appendix:metrics}. While these measures provide a structured approximation of divergent idea generation, they capture only the generative aspect of creativity and should be interpreted as approximations of structural divergence rather than comprehensive measures of creativity, which also involve evaluation, refinement, and contextual relevance.

\section{Results}
\subsection{Experimental setup}
\label{sec:experiment}
The models used for evaluation include Meta-Llama-3.1-70B-Instruct, Qwen3-30B-A3B-Instruct-2507, Gemini-2.5-Flash,  GPT-4o, and Grok-4.1-fast-reasoning. Since higher temperatures are generally associated with increased diversity and potentially more creative outputs \citep{peeperkorn2024temperature}, we set each model's temperature to the value recommended by the vendor for creative outputs: 0.7 for Qwen\citep{qwen3technicalreport} and Grok\citep{xAI2025Grok41}, and 1.0 for all remaining models. The maximum number of tokens was set to 8192 for all experiments. To account for stochastic variability induced by the temperature sampling, we generated three independent output iterations per model. Reported results correspond to the average across these three runs. Full hyperparameter configurations and infrastructure details are provided in Appendix~\ref{sec:app_experiment}.

\subsection{Model comparison}
Table \ref{tab:model_comparison} reports averages across all prompts, except for perplexity, for which we report medians due to extreme outliers. For instance, Llama has a median perplexity of 70.26 but a mean of 1803.32, indicating that most outputs are coherent while a small number of outliers inflate the average. While these outliers affect perplexity, their impact on cosine metrics is limited due to its bounded values and our sample size, making them unlikely to explain the observed differences with other models.

\begin{table}[htbp]
\centering
\small
\setlength{\tabcolsep}{4pt}
\caption{Model evaluation metrics. Bold number highlight best performances.}
\label{tab:model_metrics}
\begin{tabular}{lcccccc}
\toprule
Model & Fluency $\uparrow$ & Diversity $\uparrow$ & Rarity (Top 5) $\uparrow$ & Novelty (Top 5) $\uparrow$ & Overlap $\uparrow$ & Perplexity $\downarrow$ \\
\midrule
Gemini    & 10.43  & 0.464 & 0.238 (0.286) & 0.463 (0.523) & \textbf{0.279} & \textbf{85.91} \\
GPT       & 12.91  & 0.431 & 0.206 (0.273) & 0.428 (0.519) & 0.261 & 95.51 \\
Grok      & \textbf{20.16} & 0.482 & 0.273 (0.365) & 0.481 (0.589) & 0.218 & 244.32 \\
Qwen      & 19.87  & 0.403 & 0.226 (0.291) & 0.509 (0.588) & 0.208 & 85.92 \\
Llama70b  & 9.35  & \textbf{0.485} & \textbf{0.294 (0.372)}  & \textbf{0.556 (0.61)} & 0.258 & 70.26 \\
\bottomrule
\end{tabular}
\label{tab:model_comparison}
\end{table}

Our results show that there is some variability in performance among the models, with Grok and Llama producing outputs that seem more novel, rare, and diverse than those of other models, confirming the value of such an evaluation framework. Exposure to very original ideas is essential compared to many conventional ones to improve human creativity  as it play on cognitive overload \citep{desdevises2025paradox}, Table \ref{tab:model_comparison} also reports the top 5 scoring ideas for rarity and novelty that truly compare the divergence of an idea compared to the set of all reference ideas. The difference between the best-performing models, Grok and Llama, becomes even more evident. 

\subsection{Briefs and attribute expansion effects}
\begin{figure}[t]
  \centering
  \begin{subfigure}[t]{0.4\linewidth}
    \centering
    \includegraphics[width=\linewidth]{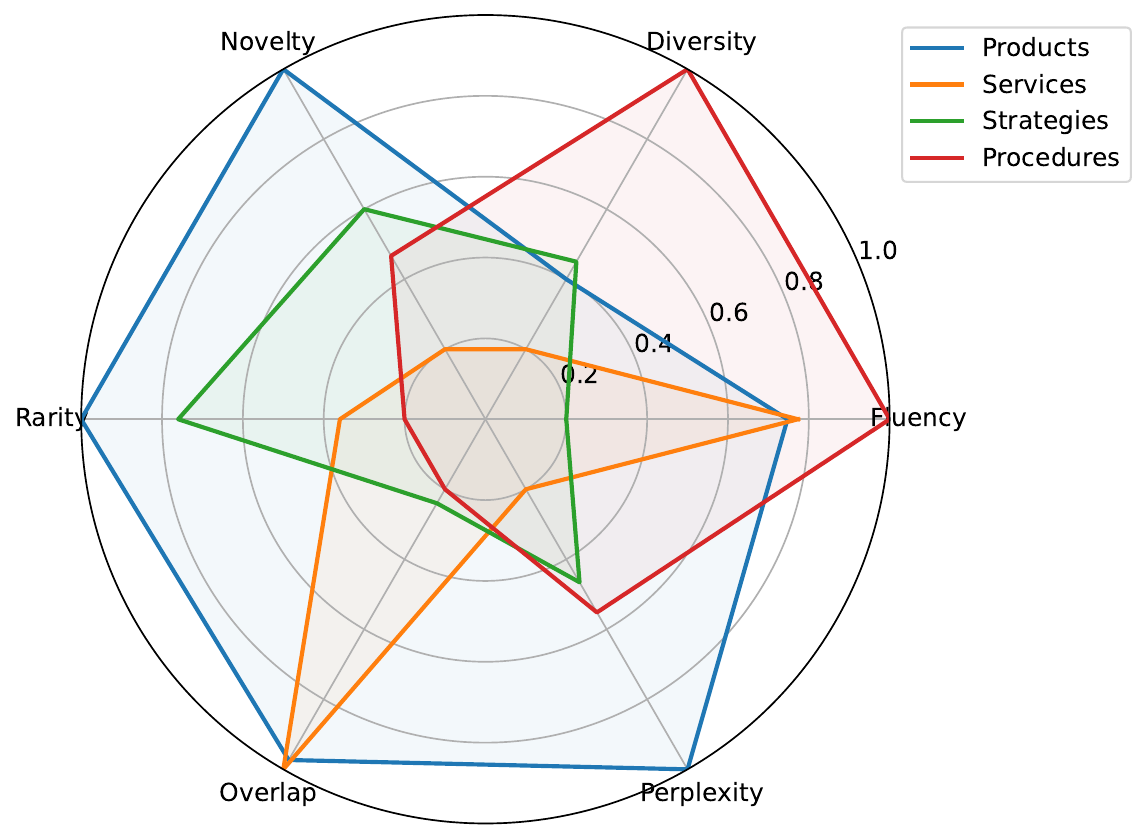}
    \caption{Performance across brief categories.}
    \label{fig:Spider_BriefType}
  \end{subfigure}
  \hfill
  \begin{subfigure}[t]{0.4\linewidth}
    \centering
    \includegraphics[width=\linewidth]{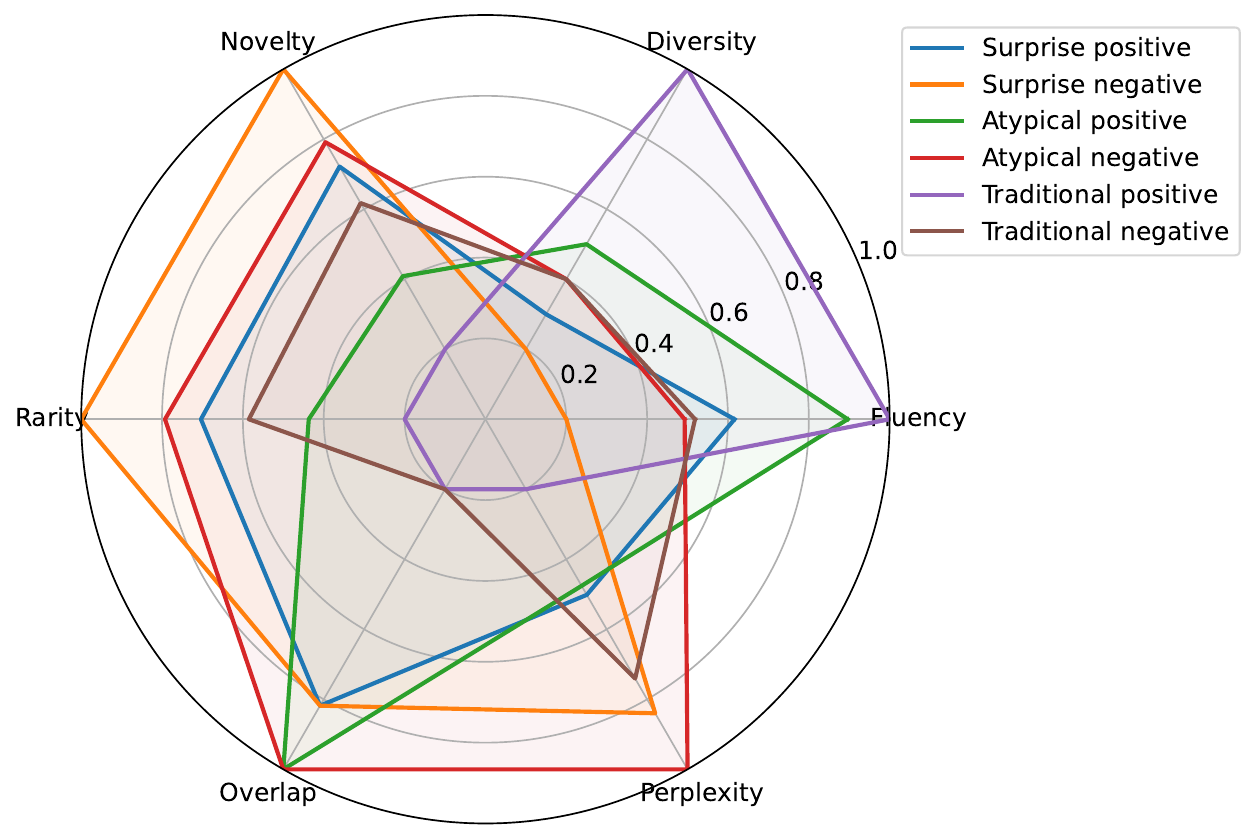}
    \caption{Performance across attribute types.}
    \label{fig:Spider_Keywords}
  \end{subfigure}
  \caption{Model performance across task and attribute dimensions.}
  \label{fig:Spider_tasks}
\end{figure}

We also analyzed the impact of the typology of brief and expansion adjectives used on the overall creative capacity of LLMs. We report the results for our six metrics for both the brief categories and the attribute properties in Figure \ref{fig:Spider_tasks}. 


Figure \ref{fig:Spider_BriefType} shows that creative performance varies by brief type. Product briefs elicit the highest novelty and rarity, suggesting that they enable more semantically unusual ideas at the expense of lower fluency and diversity. In contrast, procedure briefs produce broader and more varied outputs, while service briefs consistently scored the lowest across all metrics therefore generating the most conventional responses. All reported differences are statistically significant (Table~\ref{tab:pvalProducts}, Appendix~\ref{appendix:briefs}). These patterns suggest models reproduce certain human-like fixation patterns usually driven by mental representations of the world \citep{agogue2014impact}. Consistent with this, analyses of traditional creativity tasks, which have the highest potential to be included in LLMs training data, show that models generate more ideas but with lower novelty and diversity, indicating a bias toward familiar patterns learned during training \citep{west2025base}. Figure \ref{fig:Spider_Keywords} further supports these observations. More surprising keywords consistently increase novelty and rarity, once again at the expense of diversity, and all observed differences are statistically significant (Tables~\ref{tab:pvalNone}, \ref{tab:pvalSurprTrad}). Negative keywords further amplify these effects (Table~\ref{tab:pvalPosNeg}), with surprising negative combinations producing the most novel and rare outputs. These findings align with previous work on AUT and TTCT tasks, showing that LLMs perform well in creating connections between distant semantic combinations \citep{hubert2024current}. The increased originality observed with negative keywords is also consistent with their lower typicality in problem solving contexts, as creativity is generally framed as a positive activity \citep{cropley2008malevolent}.

Taken together, these results show that the brief type and the choice of attributes can significantly affect the originality of LLM-generated solutions. This finding resonates with developments in understanding creative processes, particularly regarding problem framing \citep{magistretti2025creative} and its influence on human creativity.

\subsection{Prompting strategies}
\begin{figure}[t]
  \centering
  \begin{subfigure}[t]{0.36\linewidth}
    \centering
    \includegraphics[width=\linewidth]{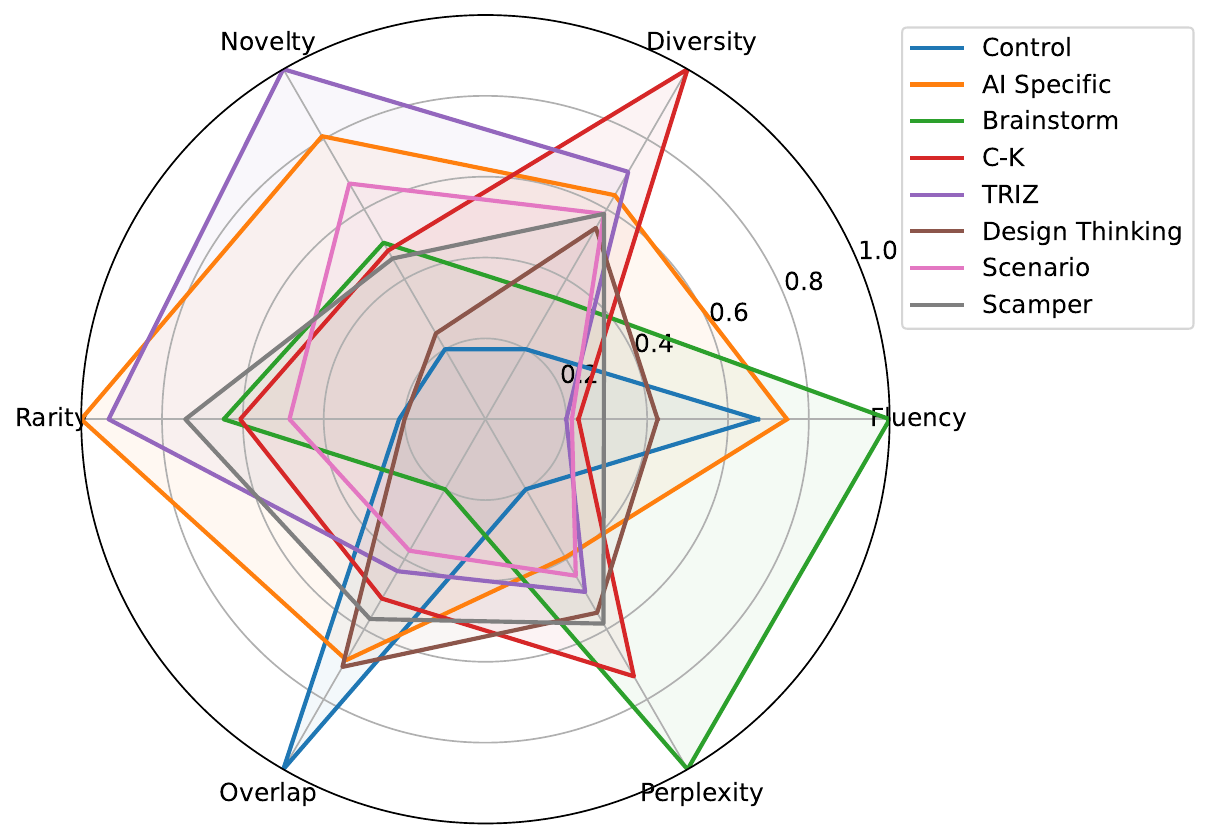}
    \caption{Prompting strategy performances.}
    \label{fig:Spider_PromptPerf}
  \end{subfigure}
  \hfill
  \begin{subfigure}[t]{0.5\linewidth}
    \centering
    \includegraphics[width=\linewidth]{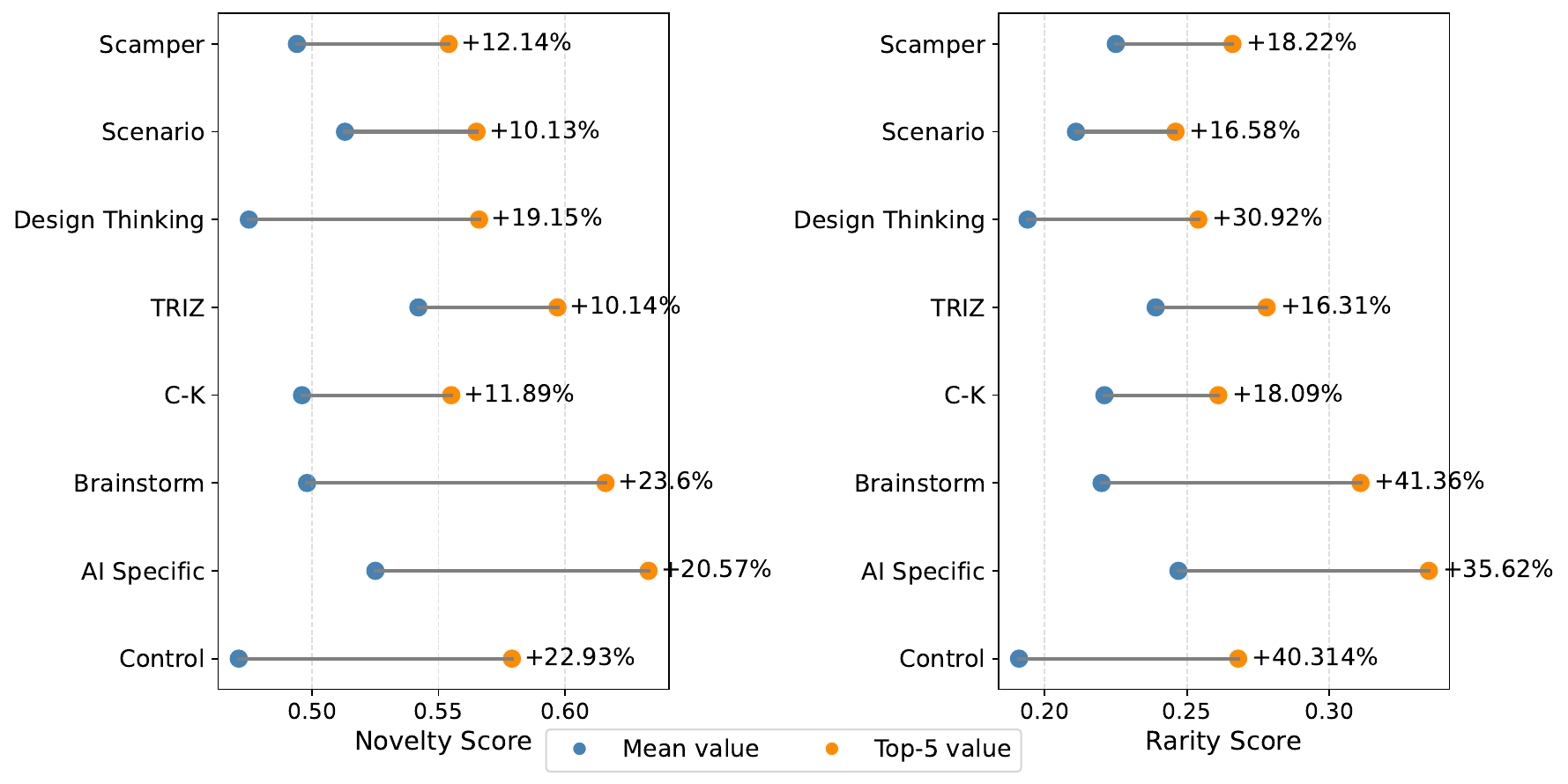}
    \caption{Increase rate of novelty and rarity for the top 5 ideas.}
    \label{fig:Spider_PromptIncrease}
  \end{subfigure}
  \caption{Effect of the best-performing alternative on model performance.}
  \label{fig:Spider_Prompts}
\end{figure}

To assess whether defixation prompting strategies could improve LLMs' creative performance, we computed the same creativity metrics across all prompting conditions. Figure \ref{fig:Spider_Prompts} reports the results for the best-performing alternative for each method, along with their performance on the Top 5 ideas for novelty and rarity. Table \ref{tab:prompt_alternatives} (in appendix \ref{appendix:prompts}) presents the aggregated results across all models for all 25 prompts tested. These results show that no single method simultaneously maximizes idea fluency, novelty, rarity and diversity. However, we observed that all method-inspired prompts consistently outperform the control prompts, and these differences are statistically significant according to t-tests (see Table \ref{tab:ttest_prompts} in Appendix \ref{appendix:prompts}) Third, in terms of novelty, two of the four alternatives developed specifically for LLMs and the third alternative of brainstorming score significantly higher on both novelty metrics, whether based on rarity or variation relative to the baseline conditions. Only the TRIZ method appears to come close to, or even exceed, their scores based on the novelty metric relative to the baseline. Moreover, the results for the top-scoring ideas show that both brainstorming and AI-specific prompts have the highest average scores for the top five ideas on the novelty and rarity metrics, and they also show the largest percentage increase compared to the average scores on these two metrics. Only the control alternative shows an equivalent increase, but with much lower initial scores on both metrics. 

Together, these results suggest that some prompting strategies do trigger higher levels of creative output from LLMs. However, prompt strategies inspired by complex methods such as C-K theory or SCAMPER do not necessarily improve the novelty of LLM-generated ideas, highlighting important differences in how creative processes operate in LLMs compared to humans. This suggests that, while human-centered methods provide useful guidance, effective prompting strategies need to be designed in alignment with the specific inductive mechanisms of LLMs. By analyzing the structure of the highest-performing prompts, we hypothesize that novelty is primarily driven by explicit instructions to "think outside the box," or keywords such as "wild" or "unconventional", joining the observation that LLMs are effective for associative creativity \citep{bellemare2024divergent}. Interestingly, simpler prompts emphasizing such cues tend to outperform more complex, method-based prompts. This suggests that gains in novelty may be more closely associated with the explicitness of divergence instructions than with the underlying structure of the creativity method.

\subsubsection{Homogeneity}
\begin{figure}[t]
  \centering
  \begin{subfigure}[t]{0.36\linewidth}
    \centering
    \includegraphics[width=\linewidth]{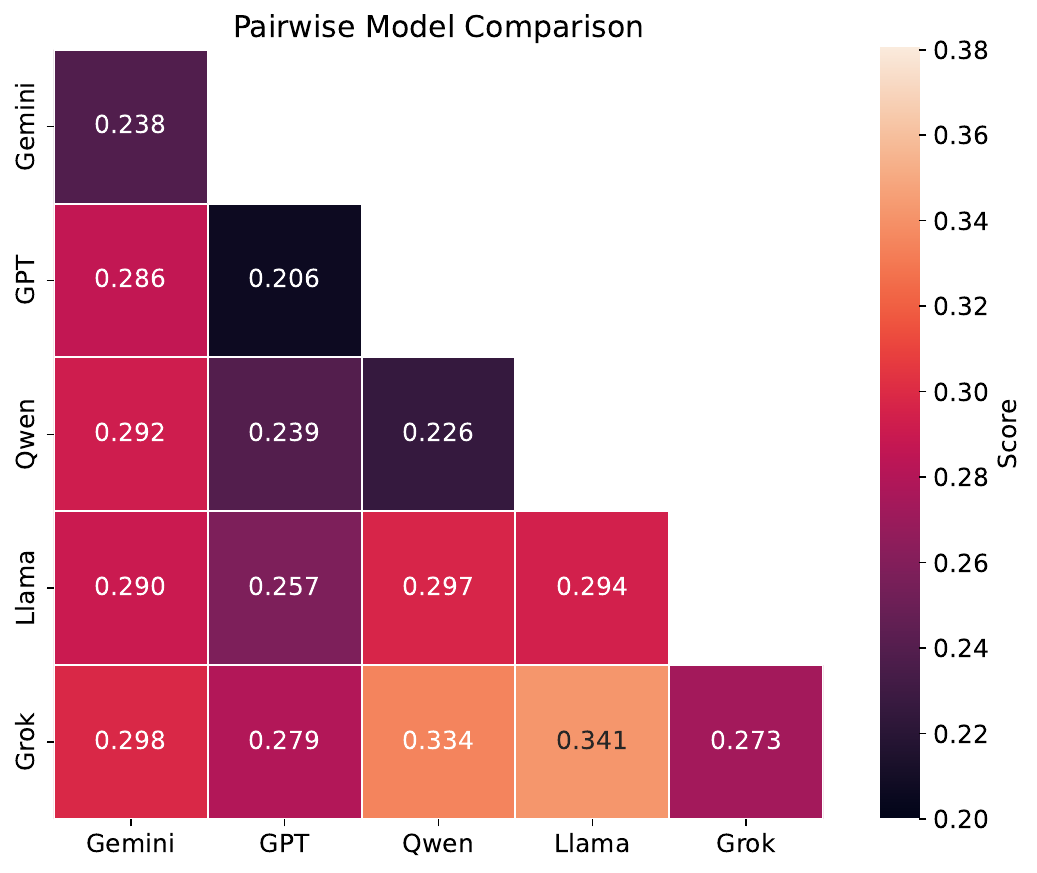}
    \caption{Distance between LLM-generated solutions.}
    \label{fig:homogen_heatmap}
  \end{subfigure}
  \hfill
  \begin{subfigure}[t]{0.39\linewidth}
    \centering
    \includegraphics[width=\linewidth]{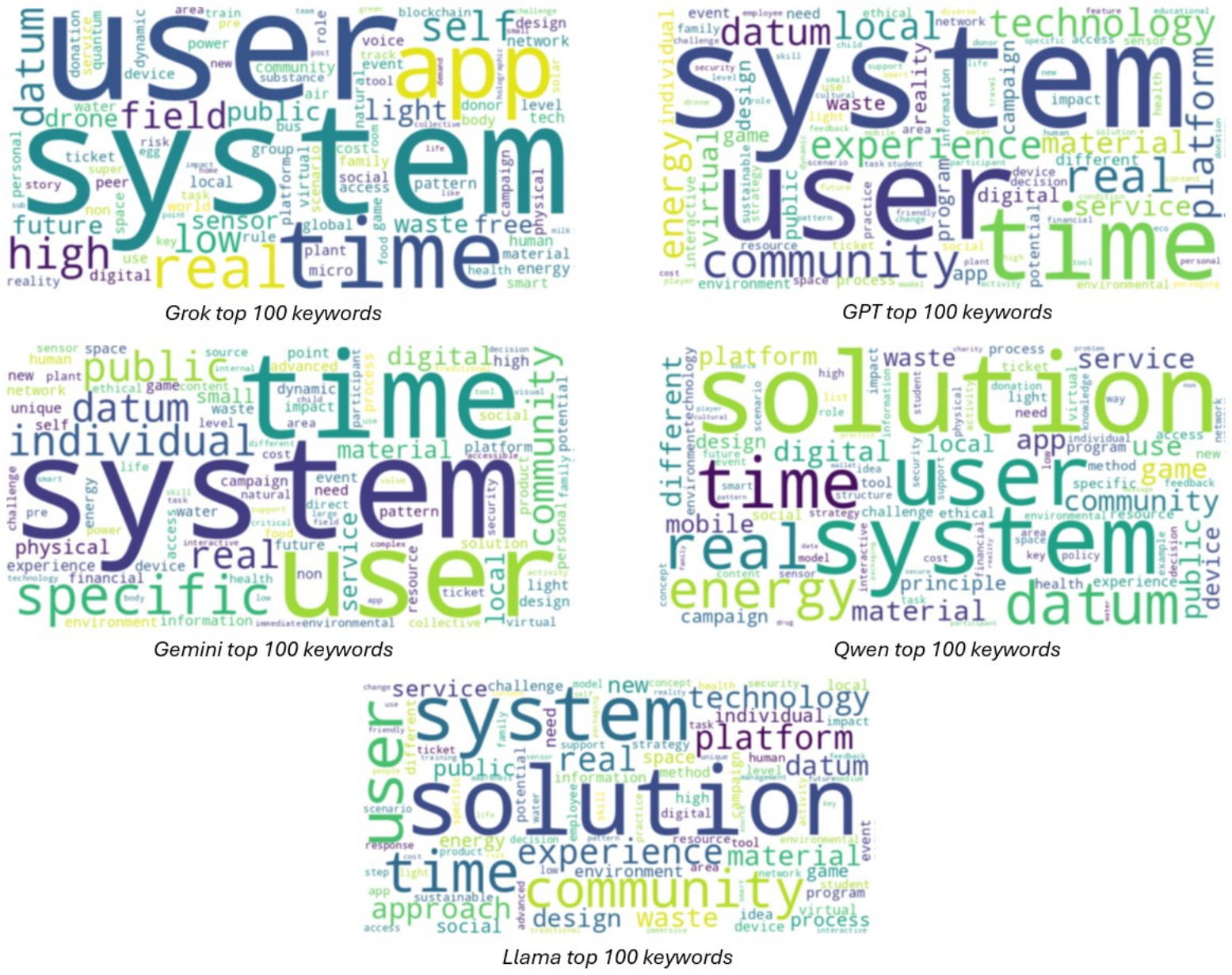}
    \caption{100 most frequent keywords for each LLM.}
    \label{fig:homogen_wordcloud}
  \end{subfigure}
  \caption{Homogenization patterns across models.}
  \label{fig:homogen}
\end{figure}

Our final objective is to assess the degree of creative divergence by analyzing whether increasing models' creativity reduces homogeneization and hivemind effects observed in LLMs \citep{jiang2025artificial}.  Figure \ref{fig:homogen_heatmap} presents the cosine distance between answers generated by all models. More specifically, we used our novelty metric, but defined our common knowledge space as the set of ideas generated by all models for a given prompt. Figure \ref{fig:homogen_wordcloud} shows the top 100 most frequently occurring keywords in the generated ideas for each model. Both graphs highlight the similarity of models' solutions, heavily oriented toward digital and technological innovation. 


Overall, these results align with the notion of an "LLM hivemind"  \citep{jiang2025artificial}, with a strong homogenization of their outputs. Our experiments also indicate that while specific briefs, keywords, and prompting strategies can increase the novelty and rarity of individual ideas, solutions proposed by LLMs remain bound within a shared semantic space, leading to persistent homogenization. This finding reinforces the hypothesis that LLMs can indeed enhance individual creative performance \citep{doshi2024generative}, but may simultaneously reduce diversity at the collective level, especially when models from various providers tend to converge toward similar outputs. Finally, these results also tend to confirm that certain forms of creativity remain difficult for LLMs to access. While they excel at recombining existing knowledge into novel variations, they appear less capable of producing transformative ideas that go beyond learned constraints \citep{franceschelli2025creativity}.

\section{Discussion}
Our results highlight the presence of fixation effects in LLM idea generation with models repeatedly converging toward familiar solution spaces. These fixation effects seem to be similar to the ones observed for humans on products or known tasks and contribute to constraining their ideation space\citep{wadinambiarachchi2024effects}. However, introducing defixating elements, such as surprising or negatively framed keywords, partially mitigates this effect, suggesting that LLM creativity is sensitive to how the ideation process is structured and can be steered beyond default responses.

While the impact of prompting strategies remains moderate, we observe consistent and statistically significant improvements in novelty when models are explicitly encouraged to “think outside the box”. In particular, as highlighted with the bold texts in figures \ref{prompt:instance_example} and \ref{prompt:instance_appendix}, the systematic use of terms like “wild, bold, surprising, unconventional, even absurd” reinforces this aspect, but what sets them apart is the explicit request to go beyond frequently used categories. Surprisingly, these prompts, and in particular the two AI-specific alternatives, are quite short and simple, breaking with the greater complexity that characterizes prompts attempting to reflect the main stages and characteristics of creativity methods. However, despite individual improvements coming from these strategies, especially for the top unconventional ideas, prompting models remain constrained in a fixation space and therefore do not solve the overall issue of output homogenization.

Finally, these results highlight that IDEAFix provides a flexible and extensible framework for studying divergent thinking capabilities of LLM, enabling controlled experiments on factors such as prompting, task structure, and contextual constraints. Its compositional design allows for virtually unlimited combinations of briefs, keywords, and prompts, facilitating the renewal of evaluation scenarios and reducing susceptibility to data contamination compared to static datasets such as AUT, DAT, or TTCT.

\section{Limitations}
\label{sec:limit}
\textbf{Dataset and prompt design.} The prompt strategies were manually designed and validated with experts, but they do not cover the full spectrum of existing defixation methods. For instance, methods such as TRIZ include a broader set of variants than those considered. In addition, adapting these methods into prompting strategies for LLMs may introduce biases, as they are primarily designed for human creativity. Finally, our study focuses on a non-iterative setting, whereas many creativity methods are typically applied through iterative, guided processes with several steps of idea refinement.

\textbf{Evaluation metrics and results.} All metrics are computed automatically, primarily relying on semantic distance in embedding space. While prior work suggests they can correlate with human judgments in open-ended tasks \citep{jiang2025artificial}, the remain imperfect proxies and should be interpreted as approximations of structural divergence rather than full measures of creativity. Future work could complement these metrics with alternative measures and human evaluations to better assess aspects such as usefulness and perceived originality.

\section{Conclusion}
We introduced IDEAFix, an evaluation framework for divergent thinking in open-ended idea generation tasks, leveraging compositional task design and structured prompting strategies. Our results show that while specific prompting strategies can increase novelty, LLM outputs remain strongly homogenized across models, revealing limits and fixation effects in their ability to explore diverse solution spaces. Overall, IDEAFix provides a scalable and extensible framework for studying LLM creativity and opens new directions for designing evaluation protocols that account for both outputs and the processes that generate them.

\FloatBarrier
\bibliography{references}
\bibliographystyle{apalike}

\FloatBarrier
\appendix

\section{IDEAFix Dataset Card Overview}
\label{appendix:dataset}

\subsection{Dataset Overview}
\label{sec:app_datacard}
IDEAFix is a compositional evaluation dataset for studying divergent thinking in large language models. It consists of 81 design briefs expanded into 567 task variations through structured attribute combinations, paired with 25 prompting strategies inspired by creativity and defixation methods. This results in a total of 14,350 input prompts. Inference outputs from multiple LLMs are provided to enable comparative analysis.

\subsection{Dataset Structure}

\begin{table}[h]
\centering
\small
\caption{Overview of the IDEAFix dataset components.}
\begin{tabular}{ll}
\toprule
\textbf{Component} & \textbf{Description} \\
\midrule
Briefs & 81 design tasks (product, service, strategy, procedure) \\
Attributes & Structured variations (typicality $\times$ sentiment) \\
Variants & 567 task configurations (Briefs  $\times$ Attributes) \\
Prompts & 25 structured prompting strategies (e.g., TRIZ, SCAMPER, C-K) \\
Inputs & 14,350 prompt instances \\
Outputs & LLM generations from multiple models \\
\bottomrule
\end{tabular}
\end{table}

Briefs define open-ended problem-solving scenarios, while attributes control task formulation through paired adjectives that vary in terms of sentiment, typicality, and lexical positioning. Prompts correspond to structured creativity methods and include multiple formulations per method to analyze the effect of prompt wording.

\subsection{Intended Use}
This dataset is designed for research purposes, specifically to evaluate and analyze the divergent thinking and creative capabilities of large language models. It is intended to support studies on how models generate diverse and original outputs, particularly across a variety of scenarios, including those in which safety constraints or refusal behaviors may be triggered. Appropriate uses include evaluating model creativity (divergent thinking), comparing prompting strategies and other attributes, and analyzing trade-offs between creativity, safety, and robustness. The dataset may also be used to study model failure modes and behavior in restricted or safety-critical domains, where prompts are explicitly included to evaluate compliance with expected safeguards. This dataset is intended for use by researchers and practitioners in natural language processing, AI safety, and related fields.

\textbf{Out-of-scope uses.} The dataset is not intended for direct training or fine-tuning of models. It should not be used to optimize systems for bypassing safety mechanisms, to identify jailbreak strategies, or to generate harmful content in real-world applications. All restricted-domain prompts are included solely for evaluation purposes, and the dataset is designed to support responsible research on model capabilities rather than to enable harmful use cases.

\subsection{Inference Setup}

We provide inference outputs from 5 LLMs (GPT-4o, Gemini-2.5-flasg, Meta-LLaMA-3.1-70B-Instruct, Qwen3-30B-A3B, Grok-4.1-Fast), comprising generated ideas and associated metadata such as prompt configuration, brief attributes, and token usage. Each configuration is evaluated across 3 runs to ensure robustness.

\subsection{Expert Involvement}
Attribute selection and prompt validation were conducted with domain experts in ethics and innovation management. Candidate attribute variations were generated and filtered through expert judgment. Prompting strategies were further reviewed by academic researchers in creativity methods (including C-K theory, brainstorming, design thinking, TRIZ, and SCAMPER), who provided structured feedback that was integrated into the final prompt formulations.

\subsection{Sensitive Content}
A subset of briefs targets sensitive or restricted domains (e.g., fraud, political persuasion, legal advice) to evaluate model creativity when it is supposed to trigger refusal behavior. These cases are explicitly flagged through the dataset's category system.

\subsection{Risks and Potential Misuse}
A potential risk of this work is that the dataset and evaluation framework could be misused to improve LLMs’ ability to generate more diverse or unconventional outputs in sensitive or restricted domains. In particular, since the dataset includes prompts related to potentially explicit or critical scenarios, it could be repurposed to study or enhance behaviors associated with policy circumvention or jailbreak strategies. To mitigate these risks, such cases are explicitly identified and included solely for evaluation purposes, and the dataset is designed to support research on model robustness, safety, and failure modes rather than to promote harmful applications.

\subsection{Availability}
The dataset, code, and inference outputs are available at an anonymous repository here: \href{https://github.com/soummyaah/CreativityAndEthicalFramework}{Code} and \href{https://huggingface.co/datasets/soummyaah/IDEAFix}{Dataset}. They will be publicly released on Hugging Face upon acceptance, and all materials are distributed under the CC BY 4.0 license.

\section{Additional Details on the Evaluation Protocol}

\subsection{Briefs design}
\label{sec:appendix_IDEAFix}

\begin{figure}[h]
    \centering
    \includegraphics[width=0.9\textwidth]{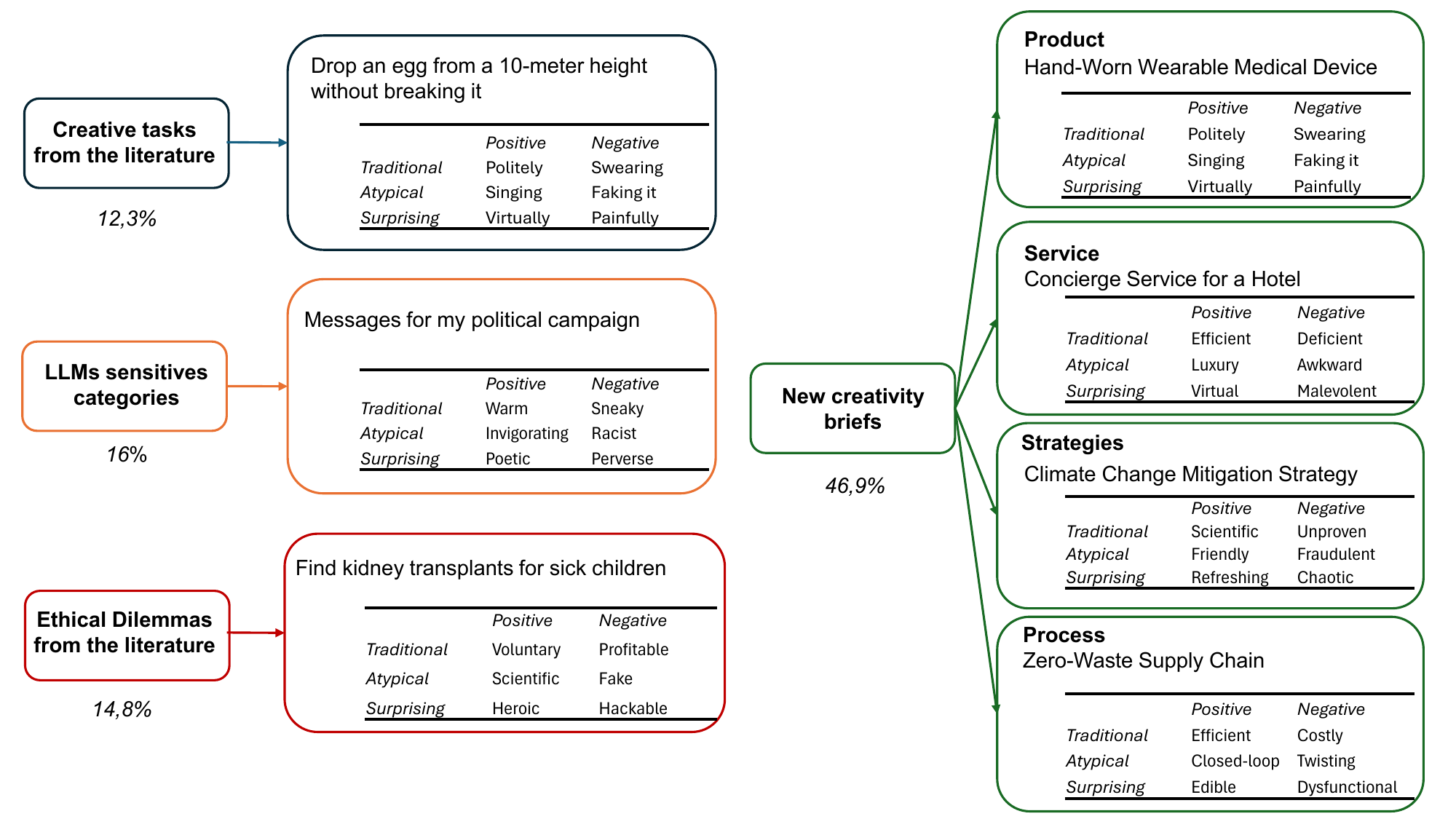}
    \caption{Examples of briefs present in IDEAFix with their associated attributes.}
    \label{fig:IDEAFix_expansion}
\end{figure}

Figure \ref{fig:IDEAFix_expansion} presents additional details of our brief expansion process with concrete example of briefs' categories (classic creative tasks, sensitive tasks, ethical dilemmas, manually designed tasks by experts) with their paired attributes; i.e: the adjectives used to complete the brief.

Moreover, a subset of the briefs in IDEAFix consists of foundational creativity tasks drawn from the social science literature on divergent thinking, creative problem-solving, and fixation effects. These tasks have been widely used to study idea generation, cognitive fixation, and creative reasoning in domains such as design, ethics, and group creativity.

Specifically, the dataset includes tasks adapted from prior work on: fixation and design theory (e.g., the “Egg Game” \citealp{agogue2014impact}), creative problem solving in groups (e.g., “Bus Use in Buffalo” \citealp{puccio2020creative}), alternative uses and incubation (e.g., “Alternative use for a brick” \citealp{gilhooly2013incubation}), organizational and ethical problem solving (e.g., “addressing employee theft” \citealp{baer2008personality}; “hand-worn wearable medical device” \citealp{mumford2010creativity}), and group ideation processes (e.g., parking solutions \citealp{garfield2001modifying}).
Additional tasks draw from studies on fixation bias and design constraints (e.g., “enter a room without a key” and “Light Up a Room Without Power Supply” \citealp{camarda2026detecting}), structured imagination and category-based generation (e.g., alien creature generation \citealp{ward1994structured}), and creativity in applied or technical domains (e.g., Mars mission design \citealp{shafirovitch2003mars}). Finally, tasks such as “Bookmark” are adapted from studies of personality and creative performance in group contexts \citealp{robert2010examination}. These tasks were selected to cover a range of cognitive processes in divergent thinking for idea generation and fixation overcoming.


\subsection{Prompts design}
\label{sec:appendix_IDEAFixPrompts}
Figure \ref{prompt:instance_appendix} presents additional examples of prompting strategies used in our dataset. The first one presents a Design thinking alternative based on evaluating stakeholders' needs, our best brainstorming alternative based on unexpectedness and explicit divergence, and our second-best-performing AI alternative that we manually created. For the last two prompting strategies, we highlight in bold textual elements that explicitly state the model should go beyond its norm and expected designs, as we hypothesize that these are key elements of the performance successes of these alternatives.

\begin{figure}
\begin{promptbox}
Design Thinking is a human-centered problem-solving method focused on understanding user needs, reframing problems, and generating innovative solutions through iterative exploration. It emphasizes empathy, creativity, and experimentation to design solutions that are desirable, feasible, and viable. The key principles include empathy for end-users, reframing challenges through insights, diverging before converging, and iterative ideation.

The method proceeds through the following steps:

1. Begin with a deep analysis of user needs using a human-centered lens. Identify all relevant stakeholders for the challenge (users, beneficiaries, decision-makers, etc.). Explore their problems, frustrations, expectations, motivations, and behaviors. If relevant, map the current user journey and identify key pain points.

2. Reframe the problem based on these insights into a clear and actionable challenge statement. Avoid jumping to technical solutions. Use a “How Might We” (HMW) format to articulate the opportunity while highlighting stakeholder impact.

3. Generate a wide range of possible solutions through divergent thinking. Draw on analogous inspiration from other industries, explore bold or extreme ideas, and iterate toward practical, viable concepts.

By following this method, its principles, and reasoning step-by-step, work on the following challenge: [Brief context].

Instructions:
List all the solutions you find as a list.
Each solution is composed of a maximum of two sentences enabling to understand the general concept behind the solution.
All solutions should be written in English.
The solutions are:
\end{promptbox}

\begin{promptbox}
Brainstorming is a creativity technique that emphasizes generating a high volume of diverse and unexpected ideas without any evaluation during the idea generation phase. The key principles include:

Focus on Quantity:
Forget about finding “the best” idea. Your only goal is volume. List as many ideas as you can — the more, the better. \textbf{Leave no stone unturned}, no matter how small or strange.

Withhold Criticism:
Avoid judging the ideas — even in your internal logic. There are no bad or wrong ideas at this stage. All contributions are valid, and \textbf{nothing is too simple, too obvious, or too ridiculous}.

Welcome Wild Ideas:
Push the boundaries. \textbf{Propose bold, unconventional, even absurd solutions.} Creativity lives in surprise. \textbf{Don’t be afraid to go far beyond what seems realistic} — breakthroughs often begin with the impossible.

Combine and Improve Ideas:
Start mixing. Build on previous ideas by combining elements, reworking them, or turning them upside down. Use ideas as seeds to grow even more varied directions.

Step-by-step process:

1. Clearly define the challenge or problem.

2. Rapidly generate as many ideas as possible without pausing to evaluate.

3. \textbf{Encourage wild, unconventional, and surprising suggestions.}

4. Record every idea without judgment or editing.

Task:
Following this method, applying its principles, and reasoning step-by-step, work on the following challenge: [Brief context].

Instructions:
List all the solutions you find as a list.
Each solution is composed of a maximum of two sentences enabling to understand the general concept behind the solution.
All solutions should be written in English.
The solutions are:
\end{promptbox}

\begin{promptbox}
Generate as many diverse and creative solutions as possible to address the following design challenge: [Brief context].
\textbf{Let's think outside the box, meaning let's go outside conventional categories of ideas.}
Instructions:
List all the solutions you find as a list.
Each solution is composed of a maximum of two sentences enabling to understand the general concept behind the solution.
All solutions should be written in English.
The solutions are:
\end{promptbox}
\caption{Examples of alternative Design Thinking, Brainstorming, and AI specific prompting strategies}
\label{prompt:instance_appendix}
\end{figure}

\subsection{Metric formulation}
\label{appendix:metrics}
In the IDEAFix dataset, each input prompt $P_i$ provided to the language models is composed of three elements: (1) a selected prompt strategy (e.g., Brainstorming Alternative 3) $P_s$, (2) a brief description $B$, and (3) its associated adjective keyword $A$, both inserted into the $[Brief context]$ placeholder of the prompt strategy, as illustrated in Figure \ref{prompt:instance_example}. Therefore, we define an input prompt as a triplet $P_i=(P_s,B,A)$. Each input prompt $P_i$ is provided to the language models to generate a list of solutions $L_i$. After applying the idea generation protocol as described in section \ref{sec:idea_splitting}, we obtained a list of ideas $L_i$ composed of N ideas $\{I_1,I_2,…,I_N\}$. For each idea, we create a set of vectorized representations $\{E_1,E_2,…,E_N\}$ using the BERT sentence transformer model \citep{reimers2019sentence}. We first create a set of 4 metrics to measure the dimensions of creative outputs. Namely, we create metrics operationalizing the notions of fluency, diversity, novelty, and rarity \citep{silvia2008assessing}. All metrics rely on the cosine distance $D$ between sets of embedding representations, as it is a common practice to compare semantic distance between concepts in social science and creativity domains \citep{feuerriegel2025using}. Here is the detail of each metric:

\textbf{Fluency:} This metric reflects the number of ideas generated for a given prompt input $P_i$. Therefore, we define fluency as $Fluency(P_i) = |L_i |$.

\textbf{Diversity:} This metric represents the average difference of individuals ideas in each list of solutions $L_i$. More specifically, we first measure how far or different an idea $I_j$ is from the other existing ideas produced in the creative session (i.e: in the current list of solutions $L_i$). The individual difference score is measured as the minimum distance with every other ideas:  $\mathrm{Difference}(I_j) = \min_{E} D(E_j, E_k)$ ,with $j \neq k$. The overall prompt diversity is then the average difference of all ideas in the set: $Diversity(P_i)=  \frac{1}{|L_i|} \sum_j \mathrm{Difference}(I_j)$.

\textbf{Novelty:} This metric represents the originality of an idea relative to a set of reference ideas we considered common or easily generated by a model. We first define the set of common ideas $L_{common}$ as the concatenation of the ideas produced by our two control prompts strategies, i.e. the neutral brief with control prompt. An idea $I_j$ is considered novel if it is distant from any idea in the set $L_{common}$. The novelty of an individual idea is then: $\mathrm{novelty_{indiv}}(I_j) = \min_{E}D(E_j, E_{common})$, where $E_{common}$ is the embedding of all ideas generated in $L_{common}$. The input prompt novelty is then: $\mathrm{novelty(P_i)} = \frac{1}{|L_i|} \sum_j \mathrm{novelty_{indiv}}(I_j)$.

\textbf{Rarity:} This metric represents the originality of an idea compared to the ensemble of ideas generated by a model for all prompting strategies $P_s$ for the same brief $(B,A)$. Therefore, rarity is defined similarly to the novelty metric, differing only in the total number of ideas considered common referent knowledge to measure it. The rarity metrics thus generalizes novelty, accounting for all 25 prompting strategies when measuring the originality of an idea.

To capture the utility of the ideas, we also measure the coherence of the generated solutions, as it is commonly accepted as a marker of value for LLM outputs \citep{sui2024confabulation}. Therefore, we designed the two following metrics to control the coherence of individual ideas: 

\textbf{Overlap:} This metric captures the extent to which the original brief and its associated keywords are represented in the final output. The brief is represented by a set of keywords $B={x_1^brief,x_2^brief,…,x_B^brief}$ We also define the set of keywords $I_j={x_1^j,x_2^j,…,x_M^j}$. Both sets are lemmatized and filtered to only preserve nouns, adjectives, and adverbs. The overlap score is then the intersection of both sets such that $\mathrm{Overlap}(I_j) = (|B\cap I_j|)/(|B|)$. The input prompt overlap is defined as the average of the individual idea overlap: $\mathrm{Overlap}(P_i) = \frac{1}{|L_i|} \sum_j \mathrm{Overlap}(I_j)$. 

\textbf{Perplexity:} Perplexity measures how well a language model predicts a given sequence. The perplexity of an idea is then measure as:$PPL(I_j) = exp(-1/(|I_j|) \sum_{(|I_j|)} \log P(x_t^j | x_1^j,\dots ,x_{t-1}^j))$. We use the open-source version of the GPT -2 model from Hugging Face\footnote {\url{https://huggingface.co/openai-community/gpt2}} as an independent evaluator of sentence perplexity for all our evaluated models. Once again, the perplexity of a prompt is then defined as: $PPL(P_i) = \frac{1}{|L_i|} \sum_j PPL(I_j)$. 

\section{Complementary Results}
\label{sec:app_results}
\subsection{Experiment and hyperparameters}
\label{sec:app_experiment}
Open-weight models were accessed via their Hugging Face implementations, while proprietary models were queried through their respective official APIs. For open-source models, we hosted the models using vLLM\footnote{\href{https://docs.vllm.ai/en/stable/}{https://docs.vllm.ai/en/stable/}}. For both open-source models(\href{https://huggingface.co/meta-llama/Llama-3.1-70B-Instruct}{Llama} and \href{https://huggingface.co/Qwen/Qwen3-30B-A3B}{Qwen}), we hosted them on 4 and 2 Quadro RTX 8000 GPUs with tensor parallelism and used 32 as the batch size. All other hyperparameters were left at default values. For API based models, default hyperparameter settings were chosen other than the values already mentioned.

\subsection{Brief analysis}
\label{appendix:briefs}
Table \ref{tab:pvalProducts} presents the p-values from t-tests comparing the mean differences observed across brief categories for each of our 6 metrics. We compared pairwise categories, demonstrating that the observed differences in novelty, rarity, and diversity are statistically significant and, therefore, that the product brief represents the category in which models perform best. The only exception comes from the perplexity score between products and other categories, such as procedures and strategies, indicating that models have similar performance levels in producing coherent outputs. 

\begin{table}[t]
\centering
\small
\setlength{\tabcolsep}{4pt}
\caption{P-values from pairwise t-tests of expansion attributes across domains}
\label{tab:pvalProducts}
\resizebox{\linewidth}{!}{%
\begin{tabular}{lcccccc}
\toprule
 & Fluency & Novelty & Diversity & Rarity & Overlap & Perplexity \\
\midrule
Product vs Services 
& $9.71{\times}10^{-45}$ 
& $3.58{\times}10^{-196}$ 
& $2.15{\times}10^{-16}$ 
& $1.60{\times}10^{-47}$ 
& $1.42{\times}10^{-24}$ 
& $1.94{\times}10^{-49}$ \\

Product vs Strategies 
& $1.64{\times}10^{-42}$ 
& $1.97{\times}10^{-228}$ 
& $1.61{\times}10^{-96}$ 
& $1.08{\times}10^{-10}$ 
& $0$ 
& $1.81{\times}10^{-2}$ \\

Product vs Procedures 
& $3.43{\times}10^{-13}$ 
& $1.11{\times}10^{-49}$ 
& $2.27{\times}10^{-204}$ 
& $1.11{\times}10^{-1}$ 
& $0$ 
& $3.95{\times}10^{-1}$ \\

Services vs Strategies 
& $3.62{\times}10^{-145}$ 
& $5.80{\times}10^{-13}$ 
& $7.80{\times}10^{-139}$ 
& $1.91{\times}10^{-89}$ 
& $0$ 
& $4.09{\times}10^{-72}$ \\

Services vs Procedures 
& $1.65{\times}10^{-70}$ 
& $0$ 
& $7.45{\times}10^{-239}$ 
& $1.42{\times}10^{-45}$ 
& $0$ 
& $1.56{\times}10^{-53}$ \\

Strategies vs Procedures 
& $3.48{\times}10^{-3}$ 
& $0$ 
& $2.46{\times}10^{-61}$ 
& $4.74{\times}10^{-4}$ 
& $1.17{\times}10^{-3}$ 
& $4.90{\times}10^{-5}$ \\
\bottomrule
\end{tabular}%
}
\end{table}

Tables \ref{tab:pvalSurprTrad}, \ref{tab:pvalNone}, and \ref{tab:pvalPosNeg} all present once again the p-value obtained after a t-test comparing the mean difference observed between this time between categories of attribute expansions. More specifically, Table \ref{tab:pvalNone} demonstrates that any attribute expansion generates a statistically significant modification of all metrics compared to generating a solution for the problem without any form of expansion. Moreover, Tables \ref{tab:pvalSurprTrad} and \ref{tab:pvalPosNeg} demonstrate that there are statistically significant differences between using surprising, traditional, or atypical keywords and using positive or negative keywords when designing the tasks to solve. 

\begin{table}[t]
\centering
\small
\setlength{\tabcolsep}{4pt}
\caption{P-values from pairwise t-tests of expansion attributes (vs. no attributes)}
\label{tab:pvalNone}
\resizebox{\linewidth}{!}{%
\begin{tabular}{lcccccc}
\toprule
 & Fluency & Novelty & Diversity & Rarity & Overlap & Perplexity \\
\midrule
Surprise 
& $2.58{\times}10^{-22}$ 
& $1.04{\times}10^{-184}$ 
& $8.93{\times}10^{-2}$ 
& $2.79{\times}10^{-76}$ 
& $0$ 
& $5.94{\times}10^{-7}$ \\

Traditional
& $3.05{\times}10^{-56}$ 
& $8.65{\times}10^{-75}$ 
& $1.11{\times}10^{-26}$ 
& $1.53{\times}10^{-58}$ 
& $0$ 
& $1.47{\times}10^{-15}$ \\

Atypical
& $4.53{\times}10^{-3}$ 
& $6.79{\times}10^{-176}$ 
& $1.12{\times}10^{-9}$ 
& $1.17{\times}10^{-19}$ 
& $2.13{\times}10^{-269}$ 
& $9.94{\times}10^{-1}$ \\
\bottomrule
\end{tabular}%
}
\end{table}

\begin{table}[t]
\centering
\caption{P-values from t-tests comparing surprising, atypical, and traditional expansion attributes.}
\label{tab:pvalSurprTrad}
\resizebox{\linewidth}{!}{%
\begin{tabular}{lcccccc}
\toprule
 & Fluency & Novelty & Diversity & Rarity & Overlap & Perplexity \\
\midrule
Surprise vs Traditional 
& $2.52{\times}10^{-11}$ 
& $6.86{\times}10^{-40}$ 
& $1.29{\times}10^{-49}$ 
& $1.33{\times}10^{-3}$ 
& $9.84{\times}10^{-18}$ 
& $3.72{\times}10^{-33}$ \\

Surprise vs Atypical 
& $2.88{\times}10^{-48}$ 
& $6.87{\times}10^{-1}$
& $9.21{\times}10^{-8}$ 
& $1.72{\times}10^{-28}$ 
& $2.81{\times}10^{-7}$ 
& $3.20{\times}10^{-12}$ \\

Traditional vs Atypical 
& $8.35{\times}10^{-102}$ 
& $4.61{\times}10^{-37}$ 
& $7.20{\times}10^{-89}$ 
& $1.12{\times}10^{-15}$ 
& $1.03{\times}10^{-42}$ 
& $1.73{\times}10^{-29}$ \\
\bottomrule
\end{tabular}%
}
\end{table}

\begin{table}[t]
\centering
\small
\setlength{\tabcolsep}{4pt}
\caption{P-values from t-tests of expansion attributes (positive vs. negative)}
\label{tab:pvalPosNeg}
\resizebox{\linewidth}{!}{%
\begin{tabular}{lcccccc}
\toprule
 & Fluency & Novelty & Diversity & Rarity & Overlap & Perplexity \\
\midrule
Positive vs Negative 
& $3.25{\times}10^{-31}$ 
& $5.99{\times}10^{-63}$ 
& $2.68{\times}10^{-22}$ 
& $1.05{\times}10^{-5}$ 
& $2.92{\times}10^{-3}$ 
& $5.84{\times}10^{-4}$ \\
\bottomrule
\end{tabular}%
}
\end{table}

Classic creativity tasks were identified by experts as those often used by scientists and facilitators in the social sciences and in the management of creativity literature \ref{sec:appendix_IDEAFix}. For this analysis, we isolated the briefs without attribute expansion for these specific tasks and reported the results for all 25 of our prompting strategies. Table \ref{tab:creatask} shows the results and highlights that for all metrics except fluency, models underperformed compared to any other prompts. It means that, for tasks and, therefore, their solutions which can easily be found on the web and thus in model training data, LLMs produce more ideas in number but with less novelty, diversity, and rarity. This tends to highlight a phenomenon of convergence of ideas toward a sort of data exposure. 

\begin{table}[t]
\centering
\small
\setlength{\tabcolsep}{4pt}
\caption{Evaluation metrics across models for classic literature briefs. We highlight in bold when results for these briefs outperform the average results of the model on the whole dataset.}
\label{tab:creatask}
\begin{tabular}{llcccccc}
\toprule
\multirow{2}{*}{Model} & \multirow{2}{*}{Brief Type} & \multicolumn{6}{c}{Metrics} \\
\cmidrule(lr){3-8}
 & & Fluency & Diversity & Novelty & Rarity & Overlap & Perplexity \\
\midrule
Gemini  & Literature & \textbf{10.729} & 0.444 & 0.395 & 0.205 & 0.260 & 113.388 \\
GPT     & Literature & \textbf{14.308} & 0.399 & 0.362 & 0.173 & 0.261 & 119.675 \\
Grok    & Literature & \textbf{22.485} & 0.453 & 0.414 & 0.230 & 0.195 & \textbf{272.721} \\
Qwen    & Literature & 19.868 & 0.378 & 0.458 & 0.196 & 0.203 & 158.200 \\
Llama   & Literature & 9.586 & 0.474 & 0.527 & \textbf{0.301} & \textbf{0.275} & 1139.231 \\
\bottomrule
\end{tabular}
\end{table}

\subsection{Prompt analysis}
\label{appendix:prompts}
For space and readability, we present in the paper results for only one alternative per prompting strategy, based on an overall assessment of performance across our 6 metrics. Table \ref{tab:prompt_alternatives} presents the full details of our results, showing how each prompting strategy affects the idea generation process. The variability in performance across approach alternatives (i.e., all C-K alternatives perform differently, the same for all TRIZ alternatives, etc.) suggests that models are more sensitive to key aspects of the prompts than to the core concepts underlying the various methodological approaches. 

\begin{table}[t]
\centering
\footnotesize
\setlength{\tabcolsep}{3pt}
\caption{Evaluation metrics across prompts and alternatives. Top-5 scores are shown in parentheses.}
\label{tab:prompt_alternatives}
\resizebox{\linewidth}{!}{%
\begin{tabular}{llcccccc}
\toprule
\multirow{2}{*}{Prompt} & \multirow{2}{*}{Alt.} 
& \multicolumn{6}{c}{Metrics} \\
\cmidrule(lr){3-8}
 & & Flu. & Div. & Novelty (Top5) & Rarity (Top10) & Overlap & PPL \\
\midrule

Control 
& Alt. 1 & 24.244 & 0.433 & 0.468 (0.586) & 0.237 (0.339) & 0.244 & 274.718 \\
& Alt. 2 & 19.231 & 0.420 & 0.474 (0.571) & 0.235 (0.318) & 0.266 & 276.275 \\

Brainstorming
& Alt. 1 & 22.895 & 0.437 & 0.483 (0.597) & 0.244 (0.340) & 0.219 & 463.828 \\
& Alt. 2 & 15.032 & 0.448 & 0.476 (0.565) & 0.235 (0.306) & 0.232 & 354.728 \\
& Alt. 3 & 30.285 & 0.438 & 0.498 (0.616) & 0.268 (0.370) & 0.214 & 738.274 \\

C-K
& Alt. 1 & 10.093 & 0.487 & 0.496 (0.555) & 0.265 (0.312) & 0.230 & 584.589 \\
& Alt. 2 & 10.684 & 0.455 & 0.472 (0.533) & 0.252 (0.299) & 0.282 & 494.255 \\
& Alt. 3 & 11.063 & 0.458 & 0.481 (0.545) & 0.252 (0.301) & 0.253 & 460.973 \\
& Alt. 4 & 10.821 & 0.465 & 0.482 (0.545) & 0.247 (0.297) & 0.239 & 538.358 \\

TRIZ
& Alt. 1 & 11.021 & 0.461 & 0.499 (0.563) & 0.257 (0.307) & 0.207 & 390.260 \\
& Alt. 2 & 9.300 & 0.465 & 0.542 (0.597) & 0.289 (0.333) & 0.226 & 445.127 \\
& Alt. 3 & 11.486 & 0.466 & 0.506 (0.576) & 0.268 (0.325) & 0.235 & 652.038 \\
& Alt. 4 & 12.431 & 0.446 & 0.490 (0.571) & 0.252 (0.313) & 0.271 & 527.240 \\

Design Thinking
& Alt. 1 & 13.874 & 0.448 & 0.464 (0.551) & 0.226 (0.294) & 0.246 & 462.703 \\
& Alt. 2 & 14.073 & 0.455 & 0.477 (0.557) & 0.245 (0.309) & 0.252 & 592.128 \\
& Alt. 3 & 14.740 & 0.446 & 0.473 (0.561) & 0.236 (0.306) & 0.241 & 462.946 \\
& Alt. 4 & 15.226 & 0.453 & 0.475 (0.566) & 0.235 (0.306) & 0.240 & 479.471 \\

Scenario
& Alt. 1 & 10.842 & 0.456 & 0.512 (0.576) & 0.249 (0.299) & 0.214 & 438.803 \\
& Alt. 2 & 10.163 & 0.455 & 0.495 (0.549) & 0.253 (0.294) & 0.242 & 647.807 \\
& Alt. 3 & 9.662  & 0.456 & 0.513 (0.565) & 0.256 (0.296) & 0.223 & 418.393 \\

Scamper
& -- & 11.741 & 0.456 & 0.494 (0.554) & 0.275 (0.321) & 0.233 & 497.491 \\

AI specific
& Alt. 1 & 15.709 & 0.463 & 0.508 (0.582) & 0.272 (0.334) & 0.192 & 646.421 \\
& Alt. 2 & 24.130 & 0.453 & 0.499 (0.607) & 0.273 (0.371) & 0.238 & 394.300 \\
& Alt. 3 & 23.622 & 0.460 & 0.525 (0.633) & 0.294 (0.397) & 0.239 & 388.089 \\
& Alt. 4 & 13.427 & 0.459 & 0.507 (0.577) & 0.285 (0.350) & 0.279 & 307.927 \\

\bottomrule
\end{tabular}%
}
\end{table}

Finally, Table~\ref{tab:ttest_prompts} reports the results of t-tests comparing the average performance of the best-performing strategies against the two control conditions. We observe that several methods—such as Design Thinking, C-K theory, and SCAMPER—struggle to significantly outperform simple prompting formulations, yielding limited or no statistically significant improvements, particularly in diversity, novelty, and rarity; key indicators of idea originality. On the contrary, brainstorming or the AI-specific methods produce significant results, highlighting that there are specific methods to steer AI to more creative output.

\begin{table}[t]
\centering
\small
\setlength{\tabcolsep}{4pt}
\caption{P-values from pairwise t-tests of prompt strategies (vs. control condition)}
\label{tab:ttest_prompts}
\resizebox{\linewidth}{!}{%
\begin{tabular}{lcccccc}
\toprule
 & Fluency & Novelty & Diversity & Rarity & Overlap & Perplexity \\
\midrule
AI specific 
& $1.82{\times}10^{-2}$ 
& $1.16{\times}10^{-34}$ 
& $1.11{\times}10^{-124}$ 
& $1.71{\times}10^{-62}$ 
& $1.05{\times}10^{-1}$ 
& $1.08{\times}10^{-2}$ \\

Brainstorming 
& $1.55{\times}10^{-56}$ 
& $8.18{\times}10^{-2}$ 
& $7.27{\times}10^{-38}$ 
& $1.06{\times}10^{-28}$ 
& $2.05{\times}10^{-4}$ 
& $7.75{\times}10^{-3}$ \\

C-K 
& $7.42{\times}10^{-200}$ 
& $1.04{\times}10^{-1}$ 
& $1.32{\times}10^{-45}$ 
& $3.12{\times}10^{-26}$ 
& $3.03{\times}10^{-1}$ 
& $1.58{\times}10^{-1}$ \\

TRIZ 
& $2.65{\times}10^{-255}$ 
& $2.24{\times}10^{-44}$ 
& $1.94{\times}10^{-1}$ 
& $3.87{\times}10^{-20}$ 
& $5.59{\times}10^{-3}$ 
& $9.56{\times}10^{-3}$ \\

Design Thinking 
& $4.86{\times}10^{-113}$ 
& $2.16{\times}10^{-2}$ 
& $1.47{\times}10^{-1}$ 
& $1.19{\times}10^{-1}$ 
& $4.22{\times}10^{-2}$ 
& $1.54{\times}10^{-1}$ \\

Scenario 
& $3.86{\times}10^{-251}$ 
& $1.18{\times}10^{-2}$ 
& $1.26{\times}10^{-9}$ 
& $1.73{\times}10^{-15}$ 
& $1.89{\times}10^{-10}$ 
& $2.39{\times}10^{-3}$ \\

Scamper 
& $5.77{\times}10^{-166}$ 
& $2.81{\times}10^{-1}$ 
& $2.23{\times}10^{-4}$ 
& $2.88{\times}10^{-16}$ 
& $8.06{\times}10^{-2}$ 
& $4.79{\times}10^{-7}$ \\
\bottomrule
\end{tabular}%
}
\end{table}

\section{Generative AI Usage Statement}
\label{sec:app_llm}
In preparing this article, we used AI-assisted coding and writing tools (ChatGPT v5.3, Grammarly) to support sentence rephrasing, table formatting, and grammatical refinement.

\end{document}